\documentclass[10pt,conference,composocconf]{IEEEtran}
\usepackage{cite}
\ifCLASSINFOpdf
   \usepackage[pdftex]{graphicx}
  \graphicspath{{img/}}
  \DeclareGraphicsExtensions{.pdf,.jpeg,.png}
\else
\fi

\usepackage{amsmath}
\usepackage{array}
\usepackage{url}
\usepackage[utf8]{inputenc}

\usepackage{amssymb} 
\usepackage{mathtools}
\usepackage{bm}
\usepackage{color}
\usepackage{xspace}
\usepackage{microtype}
\usepackage{multicol}
\usepackage{booktabs}

\usepackage{tikz}
\usepackage{pgfplots}
\usetikzlibrary{patterns,external,pgfplots.groupplots}
\usepackage{booktabs}

\usepackage{varioref}
\usepackage[pagebackref=true,breaklinks=true,letterpaper=true,colorlinks,bookmarks=false,citecolor=green!80!black,linkcolor=red!80!black,urlcolor=blue]{hyperref}
\usepackage{cleveref}
\usepackage{subcaption}

\usepackage{csquotes}

\newcommand{\onedot}{.\xspace}
\newcommand{\etal}[1]{#1~et~al\onedot}

\newcommand{\eg}{e.\,g.,\xspace}

\newcommand{\cf}{cf\onedot}
\newcommand{\ie}{i.\,e.,\xspace}

\renewcommand{\vec}[1]{\bm{#1}}
\DeclareMathOperator*{\argmin}{argmin}
\crefname{section}{Sec.}{Sections}
\crefname{figure}{Fig.}{Figure}
\crefname{table}{Tab.}{Table}
\crefname{equation}{Equ.}{Equation}

\definecolor{faublue}{RGB}{0,51,102}

\newcommand{\map}{mAP\xspace}
\newcommand{\icdar}{Historical-WI\xspace}
\newcommand{\clamm}{CLaMM16\xspace}
\newcommand{\ours}{Cl-S\xspace}
\begin{document}
\title{Unsupervised Feature Learning for Writer Identification and Writer Retrieval}

\author{\IEEEauthorblockN{Vincent Christlein\IEEEauthorrefmark{1},
Martin Gropp\IEEEauthorrefmark{1},
Stefan Fiel\IEEEauthorrefmark{2}, and
Andreas Maier\IEEEauthorrefmark{1}}
\IEEEauthorblockA{\IEEEauthorrefmark{1}
Pattern Recognition Lab, Friedrich-Alexander-Universität Erlangen-Nürnberg,
 91058 Erlangen, Germany
 }
\IEEEauthorblockA{\IEEEauthorrefmark{2}
Computer Vision Lab, TU Wien, 1040 Vienna, Austria, \\
vincent.christlein@fau.de, martin.gropp@fau.de, fiel@caa.tuwien.ac.at, andreas.maier@fau.de
}}

\maketitle

\begin{abstract}
Deep Convolutional Neural Networks (CNN) have shown great success in supervised
classification tasks such as character classification or dating. Deep
learning methods typically need a lot of annotated training data, which is not
available in many scenarios. In these cases, traditional methods are often
better than or equivalent to deep learning methods. In this paper, we propose a
simple, yet effective, way to learn CNN activation features in an unsupervised
manner. Therefore, we train a deep residual network using surrogate classes. The
surrogate classes are created by clustering the training dataset, where each
cluster index represents one surrogate class. The activations from the
penultimate CNN layer serve as features for subsequent classification tasks. We
evaluate the feature representations on two publicly available datasets. The
focus lies on the ICDAR17 competition dataset on historical document writer
identification (\icdar). We show that the activation features trained
without supervision are superior to descriptors of state-of-the-art writer
identification methods. Additionally, we achieve comparable results in the case
of handwriting classification using the ICFHR16 competition dataset on historical
Latin script types (\clamm).
\end{abstract}


\IEEEpeerreviewmaketitle

\section{Introduction}
The analysis of historical data is typically a task for experts in history or
paleography. However, due to the digitization process of archives and libraries,
a manual analysis of a large data corpus might not be feasible anymore.
We believe that automatic methods can support people working in the field of
humanities. In this paper, we focus on the task of \emph{writer identification}
and \emph{writer retrieval}. Writer identification refers to the problem of 
assigning the correct writer for a query image by comparing it with images of
known scribal attribution. For writer retrieval, the task consists of finding
all relevant documents of a specific writer. Additionally, we evaluate our
method in a classification task to classify historical script types.

We make use of deep Convolutional Neural Networks (CNN) that are able to create powerful feature
representations~\cite{Razavian14} and are the state-of-the-art tool for image
classification since the AlexNet CNN of \etal{Krizhevsky}~\cite{Krizhevsky12ICD}
won the ImageNet competition. 
Deep-learning-based methods achieve also great performance in the field of
handwritten documents classification, \eg dating~\cite{Wahlberg16ICFHR}, word
spotting~\cite{Jaderberg14DFT}, or handwritten text
recognition~\cite{Bluche13FEC}. However, such methods typically require a lot of
labeled data for each class.
We face another problem in the case of writer identification, where the writers
of the training set are different from those of the test set in the typical used
benchmark datasets.
On top of that, current datasets have only one to five images per writer.
While a form of writer adaptation with exemplar Support Vector Machines (E-SVM) is
possible~\cite{Christlein17PR}, CNN training for each query image would be very cost-intensive. Thus, deep-learning-based methods are solely
used to create robust features~\cite{Christlein15GCPR,Fiel15CAIP,Xing16}. 
In these cases, the writers of the training set serve as surrogate
classes. In comparison to this supervised feature learning, we show that deep
activation features learned in an unsupervised manner can \textit{i)} serve as
better surrogate classes, and \textit{ii)} outperform handcrafted features from
current state-of-the-art methods.

\begin{figure}[t]
	\includegraphics[height=0.15\textheight]{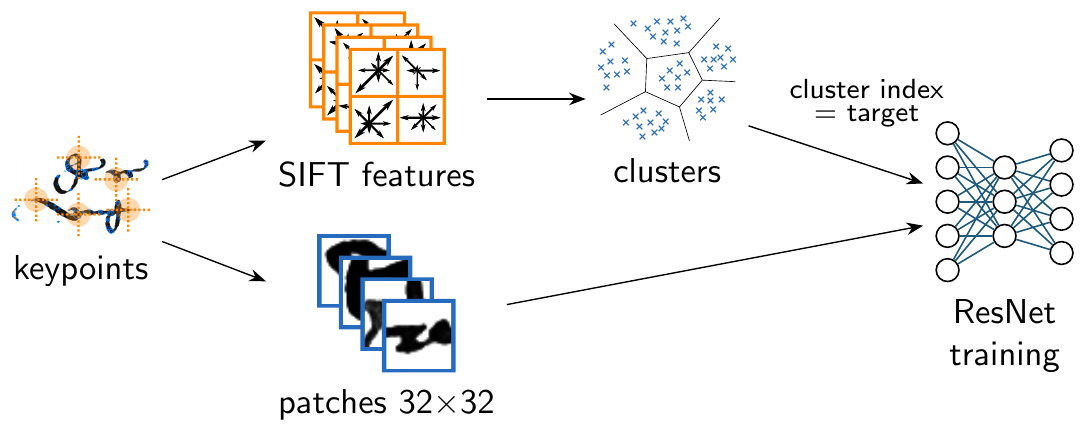}
	\caption{Overview of the unsupervised feature learning. At SIFT keypoint locations, SIFT descriptors and
	image patches are extracted. The cluster indices of the clustered SIFT
descriptors represent the targets and the corresponding patches as input for the 
CNN training.}
	\label{fig:pipeline}
\end{figure}
In detail, our contributions are as follows:
\begin{itemize}
	\item We present a simple method for feature learning using deep neural networks without
		the need of labeled data. \Cref{fig:pipeline} gives an overview of our
		method. First SIFT descriptors~\cite{Lowe04} are computed on
		the training dataset, which are subsequently clustered. 
		A deep residual network (ResNet)~\cite{He16a} is trained using patches
		extracted from each SIFT location (keypoint) using the cluster membership as target. 
		The activations of the penultimate layer serve as local feature descriptors
		that are subsequently encoded and classified. 
	\item We thoroughly evaluate all steps of our pipeline using a publicly
		available dataset on historical document writer identification. 
	\item We show that our method outperforms state-of-the-art in the case of
		writer identification and retrieval.  
	\item Additionally, we evaluate our method for the classification of medieval
		script types. On this task, we achieve equally good results as the
		competition winner. 
\end{itemize}

The rest of the paper is organized as follows. \Cref{sec:related_work} gives an
overview over the related work in the field of unsupervised feature learning,
and writer identification. 
The unsupervised feature learning and encoding step is presented in
\cref{sec:methodology}. The evaluation protocol is given in
\cref{sec:evaluation}, and the results in \cref{sec:results}.
\Cref{sec:conclusion} gives a summary and an outlook. 

\section{Related Work}
\label{sec:related_work}
We focus our evaluation on the task of writer identification and retrieval.
Method-wise, writer identification\,/\,retrieval can be divided into two groups:
\emph{statistical} methods (a.\,k.\,a.\ textural methods~\cite{Bulacu07}) and
\emph{codebook-}based methods. The differentiation lies in the creation of the
global descriptor which is going to be compared, or classified, respectively. 
Global statistics of the handwriting are computed in the former group, such as
the width of the ink trace, or the angles of stroke
directions~\cite{Brink12WIU,He14DHR}.  More recently,
\etal{Nicolaou}~\cite{Nicolaou15ICDAR} employed local binary patterns that are
evaluated densely at the image.

Conversely, codebook-based descriptors are based on the well-known
Bag-of-(Visual)-Words (BoW) principle, \ie a global descriptor is created by
encoding local descriptors using statistics obtained by a pre-trained
dictionary. Fisher vectors~\cite{Fiel13ICDAR}, VLAD~\cite{Christlein15ICDAR} or
self organizing maps~\cite{He15JDI} were employed for writer identification and
retrieval.  
Popular local descriptors for writer identification are based on Scale Invariant
Feature Transform~\cite{Lowe04} (SIFT),
see~\cite{Christlein14WACV,Christlein17PR,Fiel13ICDAR,Jain14CLF}. However, also
handcrafted descriptors are developed that are specifically designed to work
well on handwriting. One example is the work by \etal{He}~\cite{He15JDI}, who characterize script
by computing junctions of the handwriting. In contrast, the hereinafter
presented work learns the descriptors using a deep CNN. In previous works the
writers of the training datasets have been used as targets for the CNN
training~\cite{Christlein15GCPR,Fiel15CAIP,Xing16}. While
the output neurons of the last layer were aggregated using sum-pooling by
Xing and Qiao~\cite{Xing16}, the activation features of the penultimate layer
were encoded using Fisher vectors~\cite{Fiel15CAIP} and GMM
supervectors~\cite{Christlein15GCPR}. In contrast, we do not rely on any writer label
information, but use cluster membership of image patches as surrogate targets. 

Clustering has also been used to create unsupervised attributes for historical
document dating in the work of \etal{He}~\cite{He16DAS}. However, they use
handcrafted features in conjunction with SVMs.
Instead, we learn the features in an unsupervised manner using a deep CNN. 

The most closely related work comes from
\etal{Dosovitskiy}~\cite{Dosovitskiy16}, where surrogate classes are created by
a variety of image transformations such as rotation or scale. Using these
classes to train a CNN, they generate features, which are invariant to many
transformations and are advantageous in comparison to handcrafted features.
They also suggest to cluster the images in advance to apply their
transformations on each cluster image, and then use the cluster indices as
surrogate classes.
A similar procedure is applied by \etal{Huang}~\cite{Huang16ULD} to discover
shared attributes and visual representations.  In comparison to the datasets
used in the evaluation of \etal{Dosovitskiy} and \etal{Huang}, we have much more
training samples available since we consider small handwriting patches. Thus, an
exhaustive augmentation of the dataset is not necessary; instead, one cluster
directly represents a surrogate class.  
Another interesting approach for deep unsupervised feature learning is the work
of \etal{Paulin}~\cite{Paulin16}, where Convolutional Kernel Networks (CKN) are
employed. CKNs are similar to CNNs but are trained layer-wise to approximate a
particular non-linear kernel.

\section{Methodology}
\label{sec:methodology}
Our goal is to learn robust local features in an unsupervised manner. These
features can then be used for subsequent classification tasks such as writer
identification or script type classification. Therefore, a state-of-the-art CNN
architecture is employed to train a powerful patch representation using cluster
memberships as targets. A global image descriptor is created by means of VLAD
encoding.

\subsection{Unsupervised Feature Learning}
\label{sec:unsupervised}

First, SIFT keypoints are extracted. At each keypoint location a SIFT descriptor
and a $32\times32$ patch is extracted. The SIFT descriptors of the training set
are clustered. While the patches are the inputs for the CNN training, the
cluster memberships of the corresponding SIFT descriptors are used as targets.
Cf.\ also \cref{fig:pipeline} for an overview of the feature learning process.  

\begin{figure}[t]
  \includegraphics[height=0.15\textheight]{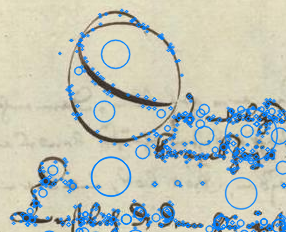}
  \hfill
	\includegraphics[height=0.15\textheight]{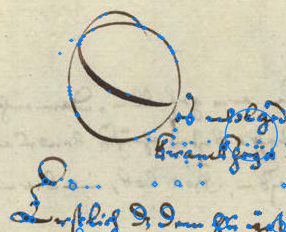}
  \caption{Excerpt of an image of the \icdar dataset. Left: Original SIFT
	keypoints, right: restricted SIFT keypoints. }
  \label{fig:sift}
\end{figure}
SIFT keypoint localization is based on blob detection~\cite{Lowe04}. The keypoints rely on
finding both minima and maxima in the Difference-of-Gaussian (DoG) scale
space, and in addition to document coordinates also contain
information about rotation and \enquote{size}, \ie their location in scale
space. The keypoints commonly occur between text lines, as can be seen in
\cref{fig:sift}.
These gratuitous locations can be filtered out either afterwards by analyzing 
the keypoint size or using the binarized image as mask.
Another possibility is to restrict the SIFT keypoint algorithm on
finding only minima in the scale space, thus, obtaining only dark on bright
blobs. We employ this technique to mainly obtain patches containing text, further
referred to R-SIFT (restricted SIFT). Note that we also filter keypoints
positioned at the same location to always obtain distinct input patches. 

For an improved cluster association, we also normalize the SIFT descriptors by
applying the Hellinger kernel~\cite{Arandjelovic12TTE}. In practice, the
Hellinger normalization of SIFT descriptors consists of an element-wise
application of the square root, followed by an $l_1$ normalization. This
normalization effectively helps to reduce the occurrence of visual bursts, \ie
dominating bins in the SIFT descriptor, and has been shown to
improve image recognition~\cite{Arandjelovic12TTE} and writer
identification\,/\,retrieval~\cite{Christlein17PR}. The descriptors are dimensionality-reduced
from 128 to 32 dimensions and whitened using principal component analysis (PCA)
to lower the computational cost of the clustering process. 

For clustering we use a subset of 500k randomly chosen R-SIFT descriptors of
the training set.  We use the mini-batch version of
$k$-means~\cite{Sculley10WSK} for a fast clustering. After the clustering
process, we filter out descriptors (and corresponding patches) that lie on
the border between two clusters. Therefore, the ratio $\rho$ between the 
distances of the input descriptor $\vec{x}$ to the closest cluster center 
$\vec{\mu}_1$ and to the second closest one $\vec{\mu}_2$ is computed, i.\,e.:
\begin{equation}
	\rho = \frac{\lVert\vec{x} - \vec{\mu}_1\rVert_2}{\lVert\vec{x} -
	\vec{\mu}_2\rVert_2}
\end{equation}
If this ratio is too large, the descriptor is removed. In practice, we use a
maximum allowed ratio of $0.9$.

Given the $32\times32$ image patches and their cluster memberships, a deep CNN
is trained. We employ a deep residual network~\cite{He16a} (ResNet) with
20-layers. Residual networks have shown great results in image classification
and object recognition. A ResNet consists of residual building blocks
that have two branches. One branch has two or more convolutional layers and the
other one just forwards the result of the previous layer, thus bypassing the other
branch. These building blocks help to preserve the identity and allow training
deeper models.
As the residual building block, we use the pre-resnet building
block of~\cite{He16b}. For training, we follow the architectural design and
procedure of \etal{He}~\cite{He16a} for the CIFAR10 dataset. Following previous
works~\cite{Christlein15GCPR,Fiel15CAIP,Xing16}, we use the activations of the
penultimate layer as feature descriptors. Note that typically the features of
the penultimate layer are most distinctive~\cite{Babenko15}, but
other layers are possible, too~\cite{Paulin16}. In our case, the penultimate layer is
a pooling layer that pools the filters from the previous residual block.  It
consists of 64 hidden nodes resulting in a feature descriptor dimensionality
of 64. 

\subsection{Encoding}
A global image descriptor is created by encoding the obtained CNN activation
features. We use VLAD encoding~\cite{Jegou12ALI}, which can be seen as a
non-probabilistic version of the Fisher Kernel. It encodes first order
statistics by aggregating the residuals of
local descriptors to their corresponding nearest cluster center. VLAD is a
standard encoding method, which has already been used for writer
identification~\cite{Christlein15ICDAR}. It has also successfully been used to 
encode CNN activation features for classification and retrieval
tasks~\cite{Gong14MSO,Paulin16}. 

Formally, a VLAD is constructed as follows~\cite{Jegou12ALI}.
First, a codebook
$\vec{D} = \left\{\vec{\mu}_1,\ldots,\vec{\mu}_K\right\}$
is computed from random descriptors of the training set using $k$-means with $K$ clusters. 
Every local image descriptor $\vec{x}$ of one image
is assigned to its nearest cluster center. Then, all residuals between the
cluster center and the assigned descriptors are accumulated for each cluster:
\begin{equation}
	\vec{v}_k = \sum_{\vec{x}_t:\; \mathrm{NN}(\vec{x}_t)=\vec{\mu}_k} (\vec{x}_t
	-\vec{\mu}_k)
	\;,
\end{equation}
where $\mathrm{NN}(\vec{x}_t)$ refers to the nearest neighbor of $\vec{x}_t$ in the
dictionary $\vec{D}$. The final VLAD encoding is the concatenation of all
$\vec{v}_k$:
\begin{equation}
	\vec{v} = \left(\vec{v}^\top_1, \ldots, \vec{v}^\top_K\right)^\top \;.
\end{equation}

We use \emph{power} normalization~\cite{Jegou12ALI} instead of the more recent
\emph{intra} normalization~\cite{Arandjelovic13AAV}. The former one is
preferable, since we employ keypoints for the patch extraction
instead of a dense sampling~\cite{Peng15}. In power-normalization, the
normalized vector $\vec{\hat{v}}$ follows as:
\begin{equation}
			\vec{\hat{v}}_i \coloneqq \operatorname{sign}(\vec{v}_i) 
			\vert \vec{v}_i
			\vert^\rho\qquad\forall
			i=\lbrace1,\ldots,\vert\vec{v}\rvert\rbrace,0<\rho\leq1\;,
\end{equation}
where we set $\rho$ to $0.5$. Afterwards, the vector is $l_2$-normalized.  

Similar to the work of \etal{Christlein}~\cite{Christlein15ICDAR}, multiple
codebooks are computed from different random training descriptors. For each of
these codebooks a VLAD encoding is computed. The encodings are subsequently
decorrelated and optionally dimensionality reduced by means of PCA
whitening. This step has been shown to be very beneficial for writer and image
retrieval\cite{Christlein15ICDAR,Jegou12NEA,Spyromitros14ACS}. We refer to this
approach as multiple codebook VLAD, or short m-VLAD. 

\subsection{Exemplar SVM}
Additionally, we train linear support vector machines (SVM) for each individual
query sample. Such an Exemplar SVM (E-SVM) is trained with only a single
positive sample and multiple negative samples. This method was originally
proposed for object detection~\cite{Malisiewicz11EOE}, where an ensemble of
E-SVMs is used for each object class. Conversely, E-SVMs can also be used
to adapt to a specific face image~\cite{Crosswhite16} or
writer~\cite{Christlein17PR}. 
In principle, we follow the approach of
\etal{Christlein}~\cite{Christlein17PR} and use E-SVMs at query time.
Since we know that the writers of the training set are independent from those of
the test set, an E-SVM is trained using the query VLAD encoding as positive
sample and all the training encodings as negatives. This has the effect of
computing an individual similarity for the query descriptor. 

The SVM large margin formulation with $l_2$ regularization and 
squared hinge loss $h(x)=\max(0,1-x)^2$ is defined as: 
\begin{equation} 
	\argmin_{\vec{w}} \frac{1}{2}\lVert\vec{w}\rVert^2_2 + c_p 
	h(\vec{w}^\top\vec{x_p}) + c_n \sum_{\vec{x_n}
	\in\mathcal{N}} h(-\vec{w}^\top\vec{x_n}) \,, 
\end{equation} 
where $\vec{x_p}$ is the single positive sample and $\vec{x_n}$ are the
samples of the negative training set $\mathcal{N}$.  
$c_p$ and $c_n$ are regularization parameters for
balancing the positive and negative costs. We chose to set them indirectly
proportional to the number of samples such that only one parameter $C$ needs to
be cross-validated in advance, \cf~\cite{Crosswhite16} for details. 

Unlike the work of \etal{Christlein}~\cite{Christlein17PR}, we do not rank the other images according
to the SVM score.  Instead, we use the linear SVM as feature
encoder~\cite{Zepeda15,Kobayashi15}, \ie we directly use the normalized weight
vector as our new feature representation for $\vec{x}$:
\begin{equation}
	\vec{x} \mapsto \hat{\vec{x}} = \frac{\vec{w}}{\lVert\vec{w}\rVert_2} \;.
\end{equation}
The new representations are ranked according to their cosine similarity.

\section{Evaluation Protocol}
\label{sec:evaluation}
The focus of our evaluation lies on writer identification and retrieval, where we
thoroughly explore the effects of different pipeline decisions.
Additionally, the features are employed for the 
classification of medieval handwriting. In the following subsections the
datasets, evaluation metrics and implementation details are presented. 

\subsection{Datasets}
The method proposed is evaluated on the dataset of the \enquote{ICDAR 2017
Competition on Historical Document Writer Identification}
(\icdar)~\cite{Fiel17ICDAR}.
The test set consists of 3600 document images written by 720
different writers. Each writer contributed 5 pages to the dataset, which have
been sampled equidistantly of all available documents to ensure a high variability
of the data. The documents have been written between the 13$^{th}$ and 20$^{th}$
century and contain mostly correspondences in German, Latin, and French.
The training set contains 1182 document images written by 394 writers. Again,
the number of pages per writer is equally distributed.

Additionally, the method is evaluated on a document classification task using
the dataset for the ICFHR2016 competition on the classification of medieval
handwritings in Latin script (\clamm)~\cite{Cloppet16}. It consists of 3000
images of Latin scripts scanned from handwritten books dated between 500 and
1600~CE.  The dataset is split into 2000 training and 1000 test images. The
task is to automatically classify the test images into one of twelve Latin
script types. 

\subsection{Evaluation Metrics}
To evaluate our method, we use a leave-one-image-out procedure, where each
image in the test set is used once as a query and the system has to retrieve a
ranked list of documents from the remaining images.
Ideally, the top entries of these lists would be the \emph{relevant}
documents written by the same scribe as the query image.

We use several common metrics to assess the quality of these results.
\textit{Soft Top N} (\textit{Soft-N}) examines the $N$ items ranked at the top
of a retrieved list.
A list is considered an \enquote{acceptable} result if there is \emph{at least 
one} relevant document in the top $N$ items.
The final score for this metric is then the percentage of acceptable results.
\textit{Hard Top N} (\textit{Hard-N}), by comparison, is much stricter and 
requires \emph{all} of the top $N$ items to be relevant for an acceptable 
result.

\textit{Precision at N} (\textit{p@N}) computes the percentage of relevant
documents in the top $N$ items of a result.
The numbers reported for \textit{p@N} are the means over all queries.

The \textit{average precision} (AP) measure considers the average \textit{p@N} 
over all positions $N$ of relevant documents in a result.
Taking the mean AP over all queries finally yields the
\textit{Mean Average Precision} score (mAP).

Since for $N = 1$ \textit{Hard-N}, \textit{Soft-N}, and
\textit{p@N} are equivalent, we record these scores only once as \textit{TOP-1}.

\subsection{Implementation Details}
If not stated otherwise, the standard pipeline consists of $5000$ cluster
indices as surrogate classes for $32\times32$ patches. The patches were
extracted from the binarized images in the case of the \icdar dataset,
and from the grayscale images in the case of the \clamm dataset. 
The patches are extracted around the restricted SIFT keypoints (see
\cref{sec:unsupervised}). 
We extract RootSIFT descriptors and apply a PCA for whitening and reducing 
the dimensionality to 32. These vectors are then used for the clustering step.
A deep residual network (number of layers $L=20$) is trained using stochastic 
gradient descent with an adaptive learning
rate (\ie if the error increases, the learning rate is divided by 10), a Nesterov
momentum of $0.9$ and a weight decay of $0.0001$. The training runs for a
maximum of 50 epochs, stopping early if the validation
error (20k random patches not part of the training procedure) increases.
Note that the maximum epoch number is sufficient given the large number of 
handwriting patches (480k).
The activations of the penultimate layer are used as local
descriptors. They are encoded using m-VLAD with five vocabularies. The final
descriptors are PCA-whitened and compared using the cosine distance.   

For the comparison with the state of the art, we also employ linear SVMs.
The SVM margin parameter
$C$ is cross-evaluated in the range $[10^{-5},10^4]$ using an inner 
stratified 5-fold cross-validation for script type classification. 
In the case of writer identification\,/\,retrieval a 2-fold cross-validation is
employed, \ie the training set is split into two writer-independent
parts to have more E-SVMs for the validation.

\section{Results}
\label{sec:results}
First, the use of writers as surrogate classes is evaluated, similar to the work
of \etal{Christlein}~\cite{Christlein15GCPR} and \etal{Fiel}~\cite{Fiel15CAIP}.
Afterwards, our proposed method for feature learning, different encoding
strategies and the used parameters are evaluated and
eventually compared to the state-of-the-art methods. 

\subsection{Writers as Surrogate Classes}
\begin{table}[t]
	\caption{Using writers from the \icdar training dataset as targets for the
	feature computation. The evaluation is carried out using the \icdar test set.} 
	\label{tab:results_writers}
	\centering
	\begin{tabular}{l*{6}c}
		\toprule
		& p@1 & p@2& p@3& p@4 & \map\\
		\midrule
		Writers (LeNet) & 66.22 & 57.10 & 48.71 & 41.70 & 44.89\\
		Writers (ResNet) & 67.36 & 58.38 & 49.81 & 42.85 & 46.11\\ 
		\bottomrule
	\end{tabular}
\end{table}
A natural choice for the training targets are the writers of the training set.
This has been successfully used by recent works for smaller, non-historical
benchmark datasets such as the ICDAR 2013 competition dataset for writer
identification~\cite{Christlein15GCPR,Fiel15CAIP}. Thus, we employ the same
scheme also for \icdar. On one hand, we employ the LeNet architecture used by
\etal{Christlein}~\cite{Christlein15GCPR}, \ie two subsequent blocks of a
convolutional layer, followed by a pooling layer, and a final fully connected
layer before the target layer with its 394 nodes.
On the other hand, we employ the same architecture we propose for our 
method, \ie a residual network (ResNet) with 20 layers. 

\cref{tab:results_writers} reveals that the use of writers as the surrogate
class does not work as intended. Independent of the architecture, we achieve
much worse results than a standard approach using SIFT descriptors or Zernike
moments, \cf~\cref{tab:sota_icdar}.

\subsection{Influence of the Encoding Method}

\begin{figure}[t!]
	\begin{tikzpicture}
		\begin{axis}[scale only axis, height=2cm, width=0.35\textwidth,
				xlabel={Number of clusters},
				ylabel={\map},
				xmin=1, xmax=16000,	ymin=52, ymax=76, 
				ytick={53.4,60,70},
				log basis x=10,	xmode=log,
				xtick={2,10,100,1000,5000,10000},
				xticklabels={2,10,100,1k,5k,10k},
			minor x tick num=2,  
			minor y tick num=1, 
			mark size=1.5pt]
			\addplot+[faublue, 
			] 
			table[header=false, x index=0, y index=1] 
			{n_clusters.txt};
		\end{axis}
	\end{tikzpicture}
	\caption{Evaluation of the number of surrogate classes (clusters) using the
	\icdar test data.}
	\label{fig:clusters}
\end{figure}
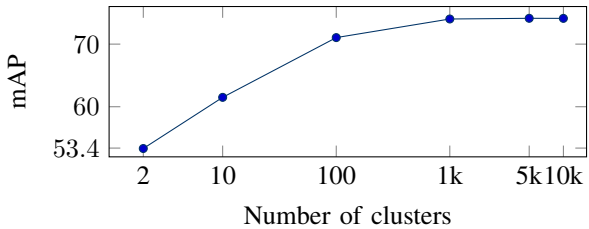

\begin{table}[t]
	\caption{Comparison of different encoding methods evaluated on the \icdar test
	test.} 
	\label{tab:parameters_enc}
		\centering
		\begin{tabular}{lcc}
			\toprule 
			& TOP-1 & \map\\
			\midrule
		\ours + Sum 							& 63.9 & 42.6\\
		\ours + FV								& 76.9 & 57.6\\
		\ours + SV							 	& 83.4 & 63.7\\
		\ours + VLAD					&	82.6 & 63.6\\
		\ours + m-VLAD						& 88.3 & 74.1\\
		\ours + m-VLAD$_{400}$ 		& 87.6 & 73.2\\
			\bottomrule
		\end{tabular}
	\end{table}
	\begin{table}[t]
		\caption{Comparison of different parameters used for the unsupervised
		feature learning step evaluated on the \icdar test set.}
	\label{tab:parameters}
		\centering
		\begin{tabular}{lcc}
			\toprule
			& TOP-1 & \map\\
			\midrule
		\ours (Baseline: $\rho=0.9$, $L=20$)	& 88.3 & 74.1\\
		\ours ($L=44$) 			& 88.2 & 74.3\\
		\ours ($\rho=1.0$) 	& 87.3 & 72.4\\
			\bottomrule
		\end{tabular}
\end{table}
	\begin{table}[t]
		\caption{Evaluation of different sampling strategies evaluated on the \icdar
		test set. Bin.\ refers to the binarized images. SIFT and R-SIFT to the SIFT
	keypoint extraction method and the restricted keypoint extraction,
	respectively, \cf \cref{sec:unsupervised}. }
		\label{tab:color}
		\centering
		\begin{tabular}{lcc}
			\toprule
			& TOP-1 & \map\\
			\midrule
			\ours (Baseline: Bin.\,/\,R-SIFT)	& 88.3 & 74.1\\
			\ours (Bin.\,/\,SIFT) 	& 88.6 & 74.8\\
			\ours (Gray\,/\,R-SIFT) & 87.1 & 71.6\\
			\ours (Gray\,/\,SIFT) 	& 87.7 & 72.3\\
			\bottomrule
		\end{tabular}
\end{table}

For the following experiments, we now train our network using the cluster 
indices as surrogate classes (denoted as \ours). 
\etal{Babenko}~\cite{Babenko15} states that sum-pooling CNN activation features 
is superior to other encoding techniques such as VLAD or Fisher Vectors. In 
\cref{tab:parameters_enc}, we compare sum-pooling to three other encoding methods: 
\textit{I)} Fisher vectors~\cite{Sanchez13} using first and second order 
statistics, which have also been employed for writer 
identification~\cite{Fiel13ICDAR}. We normalize them in a manner similar to the 
proposed VLAD normalization, \ie power normalization followed by an $L_2$ 
normalization. 
\textit{II)} GMM supervectors~\cite{Campbell06SVM}, which were used for writer 
identification by \etal{Christlein}~\cite{Christlein17PR}, normalized by a 
Kullback-Leibler normalization scheme. 
\textit{III)} the proposed VLAD encoding~\cite{Christlein15ICDAR}. 

\Cref{tab:parameters_enc} shows that sum pooling (\ours + Sum) performs significantly
worse than other encoding schemes.  While Fisher vectors (\ours + FV) trail the
GMM supervectors (\ours + SV) and VLAD encoding (\ours + VLAD / m-VLAD /
m-VLAD$_{400}$), GMM supervectors perform slightly better than the average of
the non-whitened version of the five VLAD encodings (\ours + VLAD). However,
when using the m-VLAD approach (\ours + m-VLAD), \ie jointly decorrelating the
five VLAD encodings by PCA whitening, we achieve a much higher precision. Even
if we incorporate a dimensionality reduction to 400 components (\ours +
m-VLAD$_{400}$) during the PCA whitening process, the results are significantly
better than other encoding schemes with $6400$ dimensions in case of the GMM
supervectors, or $12\,800$ in case of the Fisher vectors. 

\subsection{Parameter Evaluation}
\Cref{fig:clusters} plots the writer retrieval performance given different
numbers of surrogate classes that are used for clustering,
and the training targets, respectively. Interestingly, even a small number of
$2$ clusters is sufficient to produce better results than using the writers as
surrogate classes. When using more than 1\,000 clusters, the results are very
similar to each other with a peak at 5\,000 clusters. 

To evaluate the importance of the number of layers, we employed a much deeper
residual network consisting in total of 44 layers (instead of 20).
Since the results in \cref{tab:parameters} show that the increase in depth
(\ours ($L=44$)) produces only a slight improvement in terms of \map, and comes
with greater resource consumption, we stick to the smaller 20 layer deep 
network for the following experiments. 

Next, we evaluate the influence of the parameter $\rho$, which is used to remove
patches that do not clearly fall into one Voronoi cell computed by $k$-means,
\cf~\cref{sec:unsupervised}. When using a factor of $1.0$ (instead of $0.9$),
and thus, not removing any patches, the performance drops from $74.1\%$ \map to
$72.4\%$ \map. 

\subsection{Sampling Importance}
\label{sec:sampling_importance}
\begin{table*}[t]
	\centering
	\caption{Comparison with state-of-the-art evaluated on the \icdar test set.} 
	\label{tab:sota_icdar}
	\begin{tabular}{l*{9}cc}
		\toprule
		Method	& Top-1 & Hard-2 & Hard-3 & Hard-4 & Soft-5 & Soft-10& p@2& p@3& p@4 & \map\\
		\midrule
		SIFT + FV~\cite{Fiel13ICDAR} & 81.4 & 63.8 & 46.2 & 27.7 & 87.6 & 89.3 & 74.0 & 66.7 & 59.0 & 62.2\\
		C-Zernike + m-VLAD~\cite{Christlein15ICDAR} & 86.0 & 71.4 & 56.8 & 37.7 &
		90.3 & 91.7 & 79.9 & 73.6 & 66.4 & 69.2\\
		\midrule
		\ours 				& 88.6 & 77.1 & 64.7 & 46.8 & 92.2 & 93.4 & 83.8 & 78.9 & 72.3 & 74.8\\
		\ours + E-SVM-FE & \textbf{88.9} & \textbf{78.6} & \textbf{67.5} & \textbf{49.1} & \textbf{92.7} & \textbf{93.8} &
			\textbf{84.8} & \textbf{80.5} &	\textbf{74.0} & \textbf{76.2}\\
		\bottomrule
	\end{tabular}
\end{table*}

Finally, we also evaluate the impact of the proposed restricted SIFT keypoint
computation (R-SIFT) in comparison to standard SIFT, as well as the influence of binarization (bin.) in
comparison to grayscale patches (gray). We standardize the grayscale
patches to zero mean and unit standard deviation. \Cref{tab:color} shows
that binarization is in general beneficial for an improvement in precision.
This is even more astonishing considering that several images belong to the 
same handwritten letter.
Thus, the background information should actually improve the results. A possible
explanation could be that binary image patches are easier to train with, thus
resulting in a better representation. When comparing SIFT with its restricted version
(R-SIFT), the former consistently outperforms the restricted version by about
$0.7\%$ \map. It seems that completely blank patches do not harm the CNN
classification. This might be related to the clustering process, since all these
patches typically end up in one cluster. Furthermore, the training patches, which
are extracted, are more diverse. Also keypoints located right next to the contour
are preserved, \cf \cref{fig:sift}.

In summary, we can state that 1) m-VLAD encoding is the best encoding candidate. 
2) Our method is quite robust to the number of clusters. Given enough surrogate
classes, the method outperforms other surrogate classes that need label
information. 3) The removal of descriptors (and corresponding patches) using a simple
ratio criterion seems to be beneficial. 4) Deeper networks do not seem to be
necessary for the task of writer identification. 5) Patches extracted at SIFT
keypoint locations computed on binarized images are preferable to other
modalities. 

\subsection{Comparison with the state of the art}
We compare our method with the
state-of-the-art methods of \etal{Fiel}~\cite{Fiel13ICDAR} (SIFT + FV) and
\etal{Christlein}~\cite{Christlein15ICDAR} (C-Zernike + m-VLAD). 
While the former one uses SIFT descriptors that are encoded
using Fisher vectors~\cite{Sanchez13}, 
the latter relies on Zernike moments evaluated densely at
the contour that are subsequently encoded using the m-VLAD approach.
\Cref{tab:sota_icdar} shows that our proposed method achieves superior results
in comparison to these methods. Note that the encoding stage of the
Contour-Zernike-based method is similar to ours (\ours). It differs only in the way of
post-processing, where we use power normalization in preference to
intra normalization~\cite{Arandjelovic13AAV}. However, the difference in
accuracy is very small, see~\cite{Christlein15ICDAR}. It follows that the
improvement in performance just relies on the better feature descriptors. The
use of Exemplar SVMs for feature encoding gives another improvement of nearly $1.5$\%
\map. 

\begin{table}[t]
\centering
	\caption{Comparison with state-of-the-art evaluated on the \clamm test set.
		The numbers for the first four rows are taken from \cite{Cloppet16}.} 
		\label{tab:sota_clamm}
	\begin{tabular}{lc}	
		\toprule
	Method 		& TOP-1 \\
	\midrule
	DeepScript& 76.5\\
	FRDC-OCR 	& 79.8\\
	NNML 			& 83.8\\
	FAU 			& 83.9\\	
	\ours + SVM & \textbf{84.1}\\
	\bottomrule
	\end{tabular}
\end{table}
Additionally, we evaluate the method on the classification of medieval Latin
script types. \Cref{tab:sota_clamm} shows that our method is
slightly, but not significantly, better than state-of-the-art
methods~\cite{Cloppet16} (Soft-5: 98.1\%). Possible reasons are: a) the text
areas in the images are not segmented, \ie the images contain much more non-text
elements such as decorations, which might lower the actual feature learning
process; b) the images are not binarized, which proves beneficial, \cf
\cref{sec:sampling_importance}; c) one can train here on average with 166
instances per class, while only an exemplar classifier is trainable in the case
of writer identification. 

\section{Conclusion}
\label{sec:conclusion}
We have presented a simple method for deep feature learning using
cluster memberships as surrogate classes for local extracted image patches.
The main advantage is that no training labels are necessary. All necessary
training parameters have been evaluated thoroughly. 
We show that this approach outperforms supervised surrogate classes
and traditional features in the case of writer identification and writer
retrieval. The method achieves also comparable results to other methods on the
task of classification of script types.

As a secondary result, we found that binarized images are preferable
to grayscale versions for the training of our proposed feature learning process. 
In the future, we want to investigate this further, \eg by
evaluating only single handwritten lines instead of full paragraphs to investigate
the influence of inter-linear spaces. Activations from other layers than the
penultimate one are also worth to be examined. Another idea relates to the use of the last
neural network layer, \ie the predicted cluster membership for each patch. 
Since VLAD encoding relies on cluster memberships, this could be directly
incorporated in the pipeline. 

{\small 
\bibliographystyle{IEEEtran}
\bibliography{icdar17}
}

\end{document}